\AtBeginDocument{%
  \providecommand\BibTeX{{%
    \normalfont B\kern-0.5em{\scshape i\kern-0.25em b}\kern-0.8em\TeX}}}

\documentclass[sigconf]{acmart}
\usepackage{graphicx}  
\usepackage{subcaption}
\usepackage{multirow}
\usepackage{float}

\copyrightyear{2021}
\acmYear{2021}
\setcopyright{acmlicensed}\acmConference[MM '21]{Proceedings of the 29th ACM International Conference on Multimedia}{October 20--24, 2021}{Virtual Event, China}
\acmBooktitle{Proceedings of the 29th ACM International Conference on Multimedia (MM '21), October 20--24, 2021, Virtual Event, China}
\acmPrice{15.00}
\acmDOI{10.1145/3474085.3475348}
\acmISBN{978-1-4503-8651-7/21/10}

\begin{document}
\fancyhead{}




\title{ASFM-Net: Asymmetrical Siamese Feature Matching Network for Point Completion}

\author{Yaqi Xia}
\authornote{Both authors contributed equally to this research.}
\affiliation{%
	\institution{Xidian University}}
\email{yaqixia@stu.xidian.edu.cn}
\author{Yan Xia}
\authornotemark[1]
\affiliation{%
	\institution{Technical University of Munich}}
	\email{yan.xia@tum.de}
\author{Wei Li}
\affiliation{%
	\institution{Inceptio}}
	\email{liweimcc@gmail.com}	
\author{Rui Song}
\authornote{Corresponding author.}
\affiliation{%
	\institution{Xidian University}}
	\email{rsong@xidian.edu.cn}
\author{Kailang Cao}
\affiliation{%
	\institution{Xidian University}}
	\email{caokl_xidian@stu.xidian.edu.cn}
\author{Uwe Stilla}
\affiliation{%
	\institution{Technical University of Munich}}
	\email{stilla@tum.de}
\begin{abstract}
We tackle the problem of object completion from point clouds and propose a novel point cloud completion network employing an Asymmetrical Siamese Feature Matching strategy, termed as ASFM-Net. Specifically, the Siamese auto-encoder neural network is adopted to map the partial and complete input point cloud into a shared latent space, which can capture detailed shape prior. Then we design an iterative refinement unit to generate complete shapes with fine-grained details by integrating prior information. Experiments are conducted on the PCN dataset and the Completion3D benchmark, demonstrating the state-of-the-art performance of the proposed ASFM-Net. Our method achieves the {\bf1st place} in the leaderboard of Completion3D and outperforms existing methods with a large margin, about 12\%. The codes and trained models are released publicly at \url{https://github.com/Yan-Xia/ASFM-Net}.  
\end{abstract}




\maketitle
\begin{figure}
	\includegraphics[width=0.95\linewidth]{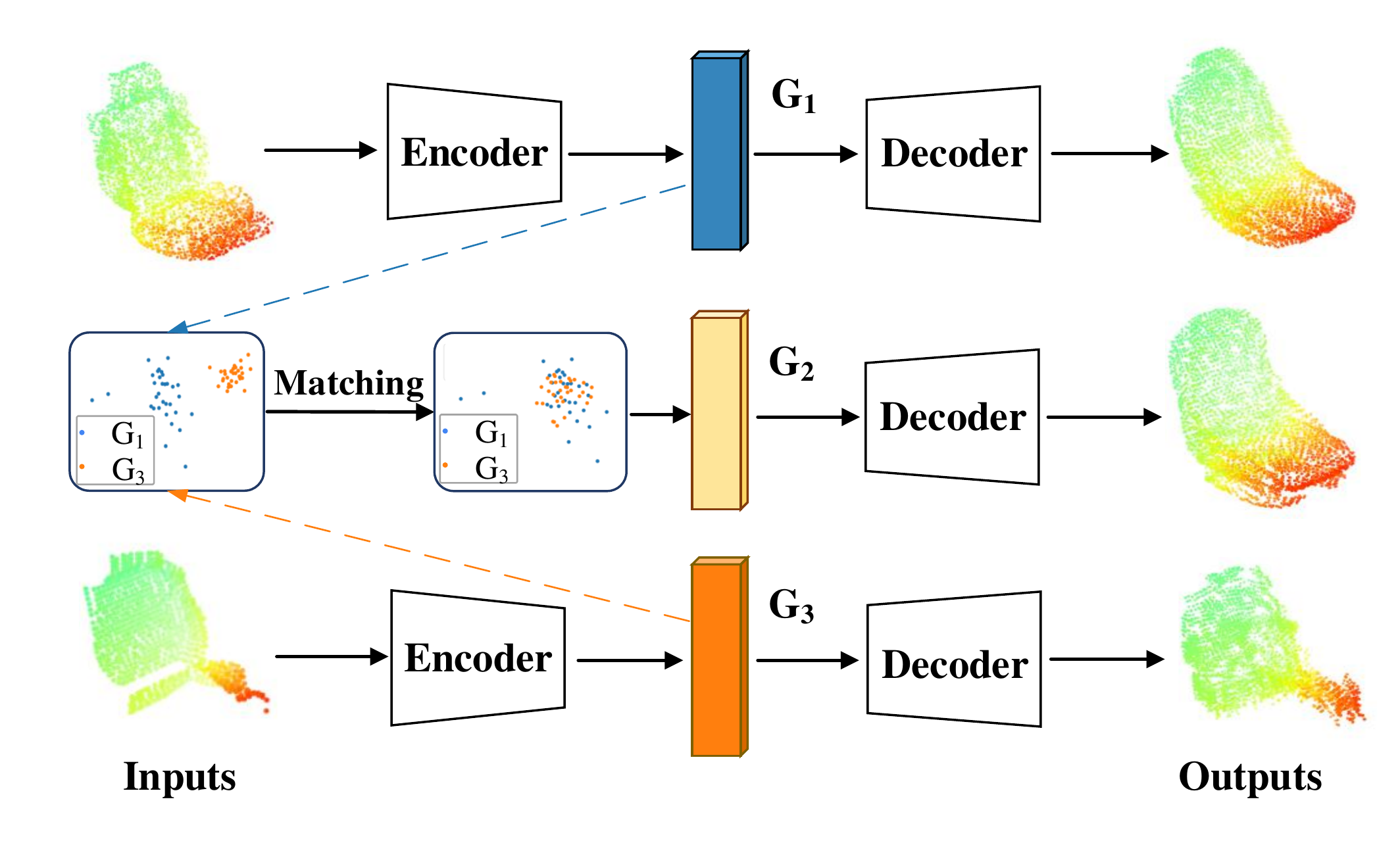}
	\caption{An illustration of our feature matching strategy. The first and the third row show the auto-encoders for the complete and partial point clouds, respectively. $G_{1}$ and $G_{3}$ represent the global features encoded from complete and partial point clouds, respectively. After feature matching, the spatial distribution between $G_{1}$ and $G_{3}$ becomes consistent ($G_{3}$ evolved into $G_{2}$).  The matched features $G_{2}$ can be used to generate the complete outputs.}
	\label{fig:sample}
\end{figure}

\section{Introduction}
Nowadays, LiDARs and depth cameras are widely applied for perceiving/scanning environments and large workspaces, which form the corner-stone of autonomous driving, VR/AR, and robotics systems. However, the scanned data are often incomplete and noisy owing to the occlusion. Shape completion methods that infer the complete structure from partial inputs have significant values in downstream applications, such as object detection~\cite{cai2018cascade, pang2019libra} and robotic manipulation~\cite{varley2017shape}.

Typical representation for 3D data includes voxelization~\cite{voxelnet,relation-shape,hypergraph}, mesh~\cite{atlasnet,graphconvolution,pixel2mesh}, multi-view images~\cite{multi-view,triplet-centerloss,rotationnet}, and point cloud~\cite{pointnet,pointnet++,pointcnn}. Among them, the point cloud is very popular due to its simple and flexible structure~\cite{xia2021vpc}. Furthermore, adding new points and interpolating them to a point cloud will be very convenient since all points in a point cloud are independent. The  unstructured property makes a point cloud easy to update~\cite{xia2019realpoint3d}, and make it difficult to apply the convolutional operation when using learning-based approaches.

PCN~\cite{pcn} firstly proposes a learning-based completion method that operates on the point clouds directly. Afterward, TopNet~\cite{topnet} adopts a novel hierarchical-structured point cloud generation decoder using a rooted tree. Furthermore, RL-GAN-Net~\cite{sarmad2019rl} uses a reinforcement learning agent to control the generative adversarial network for generating a high-fidelity completed shape. Compared to using the max-pooling operation to extract global features, SoftPoolNet~\cite{softpoolnet} proposes a soft pooling approach, which selects multiple high-scoring activation. To preserve local structures, SA-Net~\cite{sanet} uses a skip-attention mechanism to transfer local features to the decoder. However, they all rely on the global feature extracted from the partial inputs to generate complete point clouds. Recently, RFA~\cite{sfa} realizes this problem and employs a feature aggregation strategy to enhance the representation of global features. However, it still can not solve the fundamental problem: \emph{The global features only extracted from the partial inputs must be incomplete and lose the geometric details.}

To tackle this problem, we propose a novel point completion network named ASFM-Net. Specifically, we first train an asymmetrical Siamese auto-encoder network~\cite{pham2020lcd} to push the latent space extracted from the partial and complete point cloud to be as close as possible, as shown in Fig.~\ref{fig:sample}. 
In this way, the incomplete point cloud feature space can be enhanced by some shape priors, including the classification and complete geometric information. Thus, the global feature extracted by ours is more fruitful compared with previous methods. Having prior information as guidance, we can reconstruct complete point clouds with more fine-grained details. Furthermore, an iterative refinement unit is introduced to generate the final complete point clouds with the desired resolution. 

\par
The major highlights in our work are as follows:
\par
(1) We design an asymmetrical Siamese auto-encoder network to learn shape prior information, which can produce a more informative global feature for incomplete objects.
\par
(2) We propose a new point cloud completion network based on feature matching, called ASFM-Net. With the guidance of shape priors, an iterative refinement unit is introduced to retain the information of the incomplete inputs and reasonably infer the missing geometric details of the object.
\par
(3) We conduct experiments on two datasets, including the PCN dataset and the Completion3D benchmark, to demonstrate the superior performance of ASFM-Net over the previous state-of-the-art approaches. Especially, our method achieves the {\bf1st place} in the leaderboard of Completion3D, exceeding the previous state-of-the-art over 12\%.

\begin{figure*}[!t]
	\centering
	\includegraphics[width=1.0\linewidth]{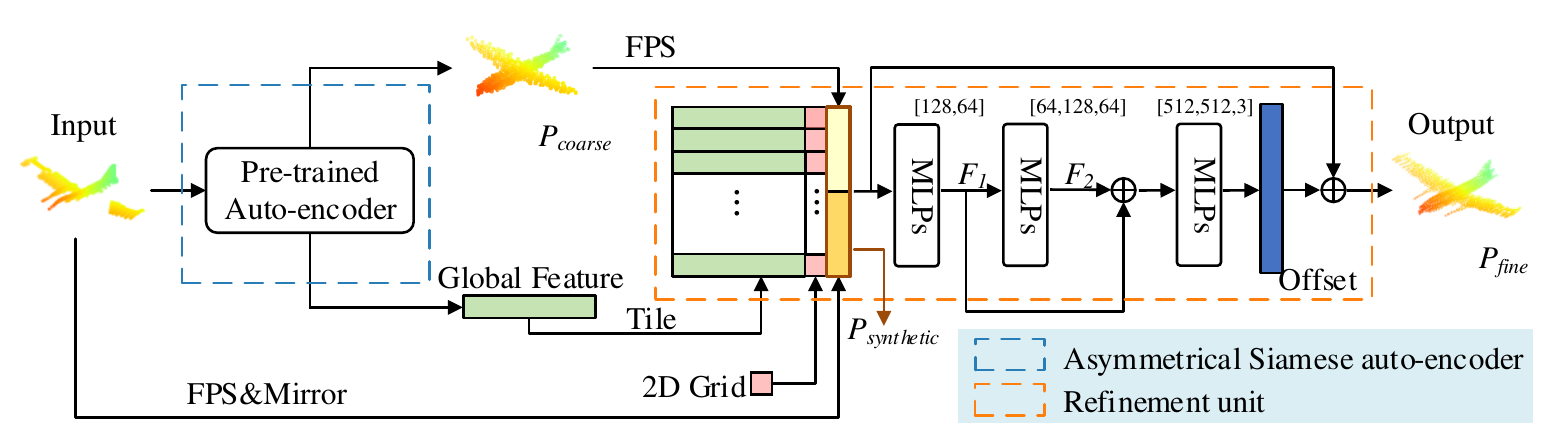}
	\caption{The overall architecture of ASFM-Net. ASFM-Net adopts a coarse-to-fine fashion to generate a dense and complete output: the asymmetrical Siamese auto-encoder module~(blue)~aims to provide a coarse point cloud and a global feature with shape prior; the refinement unit aims to preserve the details in the input and reconstruct the complete output with fine-grained geometry.}
	\label{fig:network}
\end{figure*}
\section{Related Work}
In the last few decades, 3D shape completion methods have mainly been innovated through three stages: geometry-based strategy, template-based strategy, and learning-based strategy. 
\\
\textbf{Geometry-based Approaches}
The geometry-based approaches rely on geometric attributes, such as the continuous characteristics of the surface or the symmetric properties of the space. Surface-oriented completion methods~\cite{zhao2007robust,ref1,ref2,ref3} utilize smooth interpolation to repair the defective holes on the surface of the object. The symmetric-driven methods~\cite{ref4,ref5,sung2015data} first detect the symmetric candidates of the object and then repeat the regular structure to obtain the shape information of the occluded part. Although these methods are efficient under particular circumstances, they are helpless when the missing area is large or there is no significant symmetry of the object.
\\
\textbf{Template-based Approaches}
The template-based approaches are mainly categorized into four types: direct retrieval, partial alignment, deformation, and geometric substitution. The direct retrieval methods~\cite{example,interactive,bottom} directly retrieve the closest model in the dataset as the result of completion. 
Partial alignment methods~\cite{probabilistic,data-driven,bayesian,structure,learningpart-based} divide the input object into several parts as retrieval targets and then combine them to generate the complete shape. 
The deformation methods~\cite{morphable,aligning,completing,shape} discuss the feasibility of deforming the retrieval model to fit the input shape. 
The geometric substitution methods~\cite{completion,globfit,morfit,robust} utilize the geometric primitive of the object boundary as a substitute for the retrieval target. 
However, the voluminous computational overhead makes it difficult for them to migrate to online operations. Besides, noise affects their performance significantly.
\\
\textbf{Learning-based Approaches}
With the help of the deep neural networks and the wide-ranging 3D datasets, the learning-based methods have achieved excellent performance for the shape completion task. 
Some approaches~\cite{wu20153d,voxnet,vconv} use volumetric methods to represent objects to apply 3D convolution on complex topology and tessellation learning.
PCN~\cite{pcn} first adopts a similar encoder to extract the features and outputs a dense and completed point cloud from a sparse and partial input. Furthermore, TopNet~\cite{topnet} proposes a novel hierarchical-structured point cloud generation decoder using a rooted tree. Agrawal et al.~\cite{gurumurthy2019high} adopt the GAN network to implement the latent denoising optimization algorithm. VPC-Net~\cite{xia2021vpc} is designed for vehicle point cloud completion using raw LiDAR scans.
PF-Net~\cite{pfnet} uses a multi-stage strategy to generate the 
lost structure of the object at multiple-scale. SoftPoolNet~\cite{softpoolnet} changes the max-pooling layer to a soft pooling layer, which can keep more information in multiply features. SA-Net~\cite{sanet} adopts a self-attention mechanism~\cite{xia2021soe} to effectively exploit the local structure details. Zhang et al.~\cite{sfa} propose a feature aggregation strategy to preserve the primitive details. Wang et al.~\cite{cascaded} design a cascaded refinement and add the partial input into the decoder directly for high fidelity. However, these methods use global features only extracted from partial inputs, leading to information loss in the encoding process. 


\section{Network architecture}

The overall network architecture of our ASFM-Net is shown in Fig.~\ref{fig:network}.
Given the partial input point cloud 
, we first adopt an asymmetrical Siamese auto-encoder to reconstruct the coarse point cloud $Y_{coarse}$ through unsupervised learning. It maps the partial and complete point cloud in a pre-built database into a shared latent space, with details described in Section~\ref{sec:siamese auto-encoder}. Then a refinement unit is proposed to refine $Y_{coarse}$ for producing fine-grained information, which is explained in Section~\ref{sec:refinement unit}.

\subsection{Asymmetrical Siamese Auto-encoder} \label{sec:siamese auto-encoder}

\begin{figure*}[!htbp]
	\centering
	\includegraphics[width=0.95\linewidth]{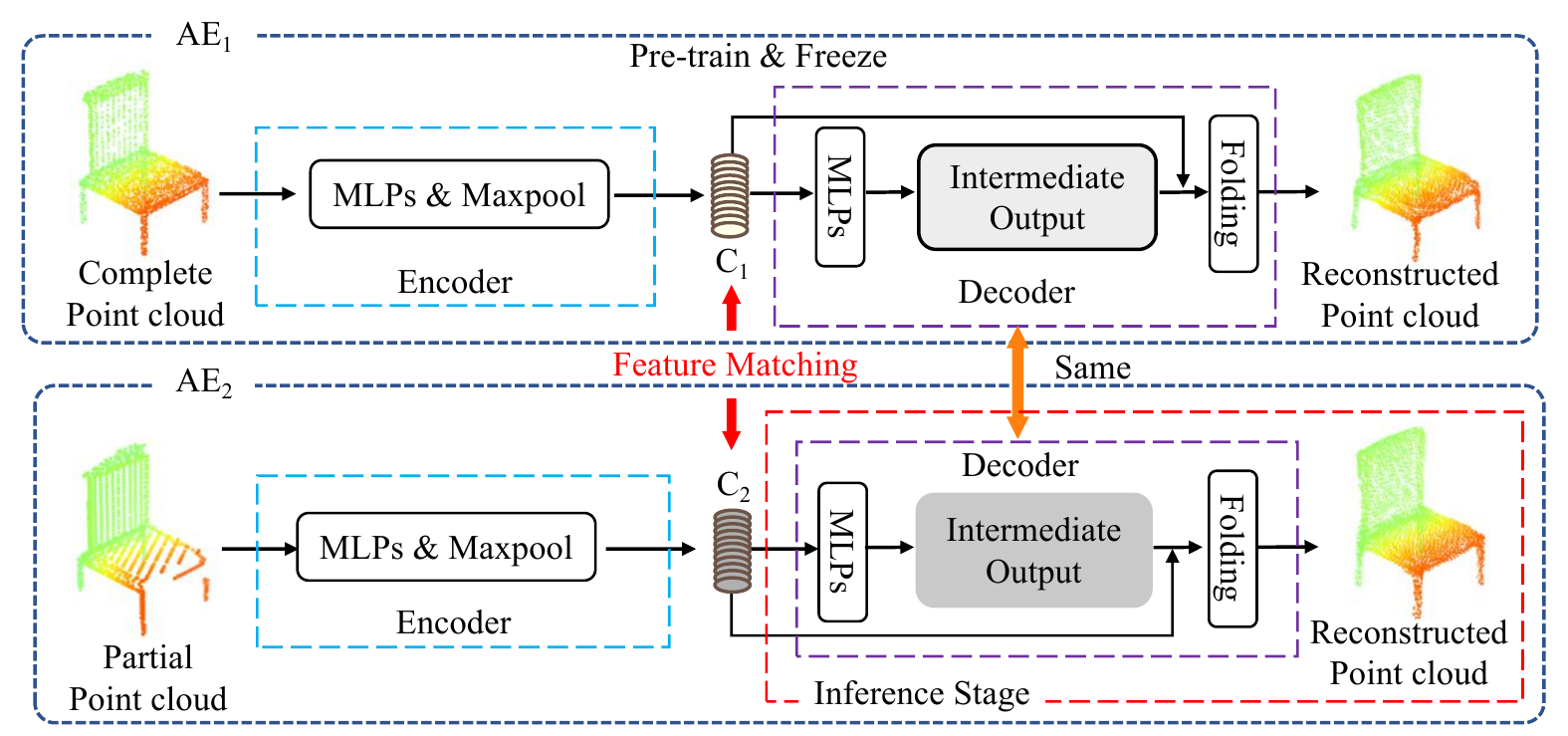}
	\caption{The network architecture of asymmetrical Siamese auto-encoder. We first train an AutoEncoder $AE_{1}$ (the upper part) and freeze the weights to produce a codeword $C_{1}$ from a complete point cloud. Then we train the encoder of another AutoEncoder $AE_{2}$ to map the partial point codeword $C_{2}$ to be consistent with $C_{1}$. In the inference stage, the decoder of $AE_{1}$ is applied to transform the $C_{2}$ to a new and complete reconstructed point cloud.}
	\label{fig:siamese auto-encoder}
\end{figure*}

To learn the global features with shape prior information from the pre-built database, we explore an asymmetrical Siamese auto-encoder network through unsupervised learning, as shown in Fig.~\ref{fig:siamese auto-encoder}. It consists of two AutoEncoder modules and a metric learning mechanism. The two AutoEncoder modules have identical architecture, and we choose PCN~\cite{pcn} as the backbone network of our auto-encoder. Any off-the-shelf point cloud feature extraction networks, e.g., PointNet~\cite{pointnet}, FoldingNet~\cite{foldingnet}, etc., can replace  PCN serving as the backbone seamlessly. However, we experimentally find PCN in our ASFM-Net achieves the best performance. 

Inspired by FoldingNet~\cite{foldingnet},  we first train an AutoEncoder $AE_{1}$ (the upper part in Fig.~\ref{fig:siamese auto-encoder}) for complete point clouds. The encoder takes each complete point cloud in the pre-built database as input and maps it to a high-dimensional codeword $C_{1}$. A decoder reconstructs point clouds using this codeword. In our experiments, the codeword length is set as 1024. Once the training process is finished, all weights of the AutoEncoder $AE_{1}$ will be frozen. Then the second AutoEncoder $AE_{2}$ (the lower part in Fig.~\ref{fig:siamese auto-encoder}) is designed for partial point clouds. The encoder also maps the partial point cloud into a 1024-dimensional codeword $C_{2}$. We hope the distribution of $C_{1}$ and $C_{2}$ will be consistent by optimizing the feature matching distance. In the inference stage, the decoder of $AE_{1}$ with fixed weights will transform the $C_{2}$ to a new and complete reconstructed point cloud. Notably, we only update the weights of the encoder in $AE_{2}$ using a feature matching loss in the training stage. Experiments in Section~\ref{sec:ablation study} demonstrate the codeword obtained using the proposed feature matching strategy is more effective than the global features directly extracted from the partial inputs.

\subsection{Refinement Unit}\label{sec:refinement unit}
Although the asymmetrical Siamese auto-encoder can extract a more effective global feature and generate a coarse point cloud $ P_{coarse} $, the fine details of the input are inevitably lost. To preserve the detailed information of the input point cloud, following ~\cite{cascaded}, we concatenate the partial inputs with the $ P_{coarse} $ to form a synthetic point cloud $ P_{synthetic} $ using the farthest points sampling (FPS) algorithm and mirror operations. We explore various symmetry operations, including plane symmetry, projective symmetry, and affine transformation operations. Experiments confirm that the XY-plane symmetry achieves the best performance. Inspired by FoldingNet~\cite{foldingnet}, we utilize a 2D grid generator and concatenates these 2D grids with each point coordinate to increase the variability of each point. In order to narrow the distribution difference between the partial and the complete point cloud, the refinement unit concatenates the global feature with the coordinate of each point in $ P_{synthetic} $. Due to the superiority of neural networks in residuals prediction~\cite{pixel2mesh}, the refinement unit predicts the coordinate offset for every point between the point set $ {{P}_{synthetic}} $ and the ground truth point cloud. Specifically, we pass the $ {{P}_{synthetic}} $ through a series of bottom-up and top-down structural styles of MLPs. 
Overall, the final completed point cloud $P_{fine}$ after refinement unit can be expressed as:
\begin{equation}
	{P_{fine}} = R\left( {{P_{synthetic}}} \right) + {P_{synthetic}}
\end{equation}
where $ R( \cdot ) $ denotes the function of predicting the coordinate residuals for the $ {{P}_{synthetic}} $. Besides, we can regard $ P_{fine} $ as the synthetic point cloud  $ {{P}_{synthetic}} $ for a new loop when a higher point resolution is required. The point resolution will be doubled by iterating the refinement operation continuously.

\subsection{Loss Function} \label{sec:loss function}
Our training loss consists of two components, a feature matching loss, and a reconstruction loss. The former requires a more similar distribution of partial and complete point clouds and the latter expects the topological distance between the completed point clouds and the ground truth as small as possible.  


{\bf Feature matching loss.} We experiment with various metrics for feature matching, such as cosine similarity~\cite{cosine}, Manhattan distance~\cite{manhattan}, and Euclidean distance~\cite{euclidean}. Finally, we choose the Euclidean distance due to the best performance. 
The similarity between two high-dimensional feature vectors can be calculated as the following equation:
\begin{equation}
	\mathcal { L}_{feat}(X,Y) = \sum\limits_{i = 0}^n {{{\left\| {F_{p_{i}} - {F_{c_{i}}}} \right\|}_2}}, 
\end{equation}
where $ X  $ and $ Y $ represent the partial and complete point cloud, respectively. $F_{p} = {({x_1},{x_2}, \cdots ,{x_n})^T}$ and $F_{c} = {({y_1},{y_2}, \cdots ,{y_n})^T}$ denote the features encoded from $ X  $ and $ Y $.

\begin{figure*}[!htbp]
	\centering
	\includegraphics[width=0.95\linewidth]{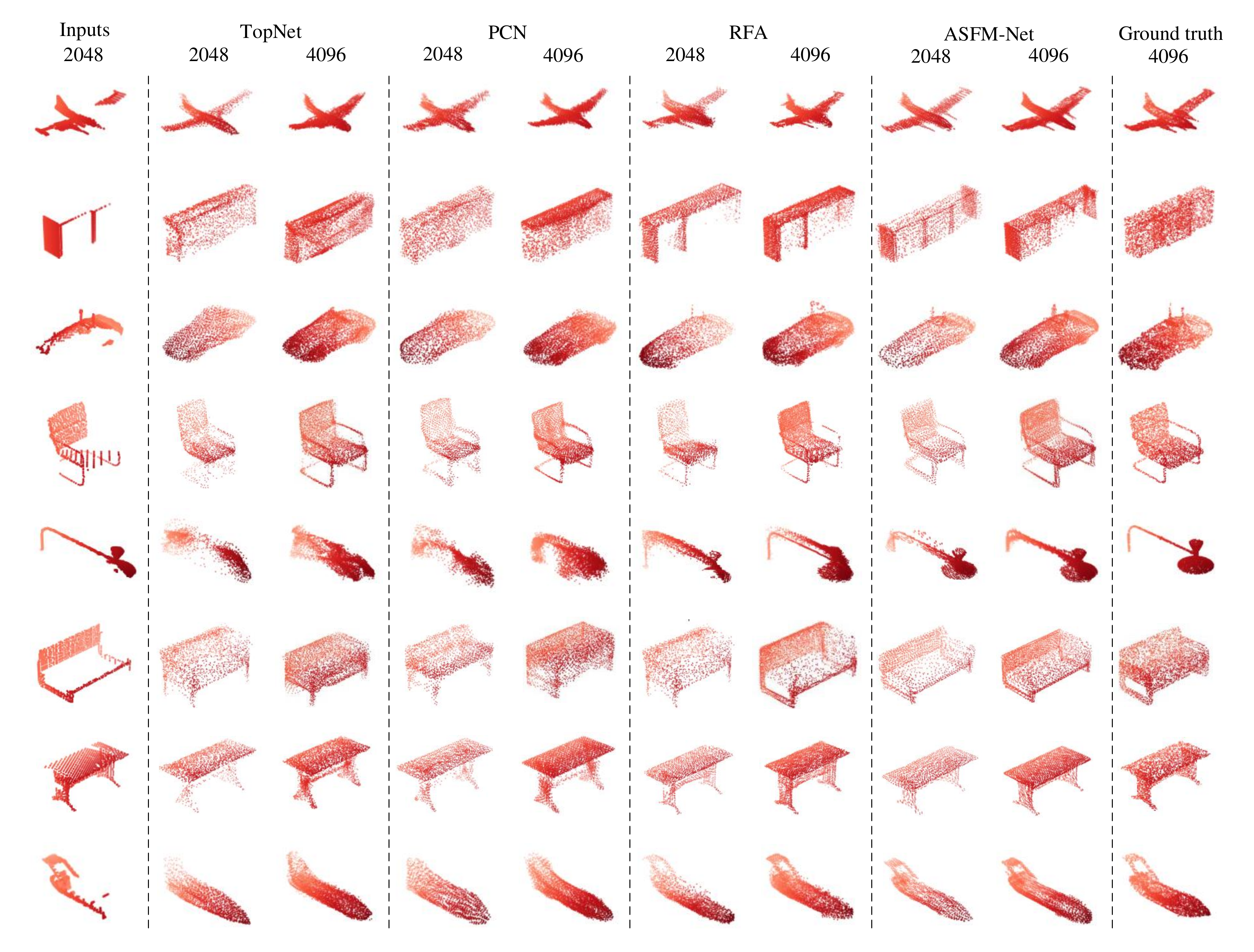}
	\caption{Qualitative comparison on known categories on the Completion3D benchmark and the PCN dataset. The 2048 and 4096 resolutions of the outputs are completed from the Completion3D benchmark and the PCN dataset, respectively.}
	\label{fig:compare}
\end{figure*}

{\bf Reconstruction loss.} Following the previous work, Chamfer Distance (CD) is used to evaluate the similarity between two sets of point clouds. There are two forms of CD: CD-T and CD-P. The definitions of CD-T and CD-P between two point clouds  ${P} $ and $ {Q} $ are as follow:

\begin{equation}
	\begin{split}
		& {\mathcal{L}_{P,Q}} = \frac{1}{{{N_P}}}\sum_{p \in P} \min _{q \in Q}\|p-q\|_{2}^{2} , \\
		& {\mathcal{L}_{Q,P}} = \frac{1}{{{N_Q}}}\sum_{q \in Q} \min _{p \in P}\|p-q\|_{2}^{2} , \\
		& {\mathcal { L}_{CD-T}}(P,Q) = {{\mathcal{L}_{P,Q}}}  +  {{\mathcal{L}_{Q,P}}} , \\
		& {\mathcal { L}_{CD-P}}(P,Q) = (\sqrt {{\mathcal{L}_{P,Q}}}  + \sqrt {{\mathcal{L}_{Q,P}}} )/2 ,
		\label{formula:3}
	\end{split}
\end{equation}
where ${N_P}$ and ${N_Q}$ are the amounts of points in ${P} $ and $ {Q} $, respectively. Notably, CD-P is used in all our experiments during the training stage.

{\bf Overall loss.} The overall loss function is the weighted sum of the feature matching loss and the reconstruction loss. Since we adopt a coarse-to-fine training fashion, both the predicted coarse point clouds $ P_{coarse} $ and final results $ P_{final} $ are optimized via the CD loss.
More formally, the overall loss is defined as:
\begin{equation}
	\begin{aligned}
		{\mathcal{L}_{sum}} = \alpha {\mathcal{L}_{feat}}( X, Y)   + \gamma {\mathcal{L}_{CD}}({P_{final}},{P_{gt}})
		\\+\beta ({\mathcal{L}_{CD}}({P_{coarse}},{P_{gt}})
		\label{formula:5}
	\end{aligned}
\end{equation}
where ${P_{gt}}$ is the ground truth point cloud. $ \alpha $, 	$ \beta $, and $ \gamma $ are all hyperparameters to balance their relationship, which are changed with the training steps synchronously.

\begin{table*}[h]
	\centering
	\renewcommand{\arraystretch}{1.0}
	\setlength{\tabcolsep}{4.0mm}
	\caption{Quantitative comparison on known categories on the PCN dataset. Point resolutions for the output and ground truth are 4096. For CD-P, lower is better.}
	\label{table:shapenet on test}
	\begin{tabular}{c|c c c c c c c c|c}
		\hline
		\multirow{2}{*}{Methods} & \multicolumn{9}{c}{Chamfer Distance($ 10^{-3} $)}                                                  \\ \cline{2-10} 
		& Airplane   & Cabinet   & Car    & Chair   & Lamp    & Sofa    & Table   & Watercraft   & Average   \\ \hline \hline
		TopNet~\cite{topnet}              & 8.21 & 15.99 & 13.28 & 15.80 & 15.29 & 17.49 & 13.27 & 13.81 & 14.14 \\
		PCN~\cite{pcn}                 & 7.95 & 15.59 & 13.10 & 15.47 & 15.31 & 16.78 & 13.22 & 13.37 & 13.85 \\
		RFA~\cite{sfa}                 & 7.49 & 15.68 & 13.52 & 14.00 & 12.33 & 16.50 & 11.99 & 11.40 & 12.87 \\ \hline
		ASFM-Net              & \textbf{6.75} & \textbf{14.85} & \textbf{12.51}  & \textbf{13.17} & \textbf{11.66} &\textbf{15.38}  &\textbf{11.49}  & \textbf{10.96} & \textbf{12.09} \\ \hline
	\end{tabular}
\end{table*}

\begin{table*}[h]
	\centering
	\renewcommand{\arraystretch}{1.0}
	\setlength{\tabcolsep}{3.5mm}
	\caption{Quantitative comparison on known categories on the Completion3D benchmark. Point resolutions for the output and ground truth are 2048. For CD-T, lower is better.}
	\label{table:c3d on c3d}
	\begin{tabular}{c|c c c c c c c c|c}
		\hline
		\multirow{2}{*}{Methods} & \multicolumn{9}{c}{Chamfer Distance($ 10^{-4} $)}                                       \\ \cline{2-10} 
		& Airplane   & Cabinet   & Car    & Chair   & Lamp    & Sofa    & Table   & Watercraft   & Average   \\ \hline \hline
		FoldingNet~\cite{foldingnet}     & 12.83         & 23.01          & 14.88         & 25.69          & 21.79          & 21.31          & 20.71          & 11.51         & 19.07          \\
		PCN~\cite{pcn}            & 9.79          & 22.70           & 12.43         & 25.14          & 22.72          & 20.26          & 20.27          & 11.73         & 18.22          \\
		PointSetVoting~\cite{pointsetvoting} & 6.88          & 21.18          & 15.78         & 22.54          & 18.78          & 28.39          & 19.96          & 11.16         & 18.18          \\
		AtlasNet~\cite{atlasnet}       & 10.36         & 23.40           & 13.40          & 24.16          & 20.24          & 20.82          & 17.52          & 11.62         & 17.77          \\
		RFA~\cite{sfa}  & 6.52 & 26.60  & 10.83 & 27.86 & 23.21 & 23.58 &11.66  & 7.41 & 17.34 \\ 
		TopNet~\cite{topnet}         & 7.32          & 18.77          & 12.88         & 19.82          & 14.60           & 16.29          & 14.89          & 8.82          & 14.25          \\
		SoftPoolNet~\cite{softpoolnet}    & 4.89          & 18.86          & 10.17         & 15.22          &12.34          & 14.87          & 11.84          & 6.48          & 11.90           \\
		SA-Net~\cite{sanet}         & 5.27          & 14.45 & 7.78 & 13.67          & 13.53          & 14.22          & 11.75          & 8.84          & 11.22         \\
		GR-Net~\cite{grnet}   & 6.13          & 16.90         & 8.27          & 12.23         & 10.22         & 14.93         & 10.08         & 5.86          & 10.64         \\
		CRN~\cite{cascaded}      & 3.38          & 13.17         & 8.31          & 10.62         & 10.00         & 12.86         & 9.16          & 5.80          & 9.21          \\
		SCRN~\cite{scrn}     & 3.35          & 12.81         & 7.78          & 9.88          & 10.12         & 12.95         & 9.77          & 6.10          & 9.13          \\
		VRC-Net~\cite{vrcnet}  & 3.94          & 10.93         & 6.44          & 9.32          & 8.32          & 11.35         & 8.60          & 5.78          & 8.12          \\
		ASFM-Net & \textbf{2.38} & \textbf{9.68} & \textbf{5.84} & \textbf{7.47} & \textbf{7.11} & \textbf{9.65} & \textbf{6.25} & \textbf{4.84} & \textbf{6.68} \\
		\hline
	\end{tabular}
\end{table*}

\section{Experiments}
\subsection{Evaluation metrics}
We compare our method with several current point cloud completion networks~\cite{foldingnet,pcn,pointsetvoting,atlasnet,sfa,topnet,softpoolnet,sanet,grnet,cascaded,scrn,vrcnet}. 
For a fair comparison, Chamfer Distance is used to evaluate quantitatively.
For the Completion3D benchmark, the online leaderboard adopts CD-T. Thus, we adopts CD-T for experiments in Section~\ref{sec:evaluation on Completion3D benchmark} and CD-P for the remaining experiments. 

%
%
%

\subsection{Evaluation on PCN dataset}\label{sec:evaluation on PCN dataset}
PCN dataset~\cite{pcn} is created from the ShapeNet dataset~\cite{shapenet}, containing pairs of complete and partial point clouds. Notably, each complete model includes 16384 points and is corresponding to eight partial point clouds. The dataset covers 30974 CAD models from 8 categories: airplane, cabinet, car, chair, lamp, sofa, table, and watercraft. Following~\cite{pcn}, the number of models for validation and testing are 100 and 150, respectively. The remaining models are used for training. In our experiments, we uniformly downsample the complete shapes from 16384 points to 4096. The performance is evaluated on the resolution containing 4096 points.

We compare against several state-of-the-art point cloud completion methods such as PCN~\cite{pcn}, TopNet~\cite{topnet} and RFA~\cite{sfa} qualitatively and quantitatively. Table~\ref{table:shapenet on test} shows that ASFM-Net achieves the lowest CD values in all eight categories, which demonstrates the superior performance of our method in this dataset. Especially, ASFM-Net improves the average CD values by $ 6.1\% $ compared to the second-best method RFA. Besides, the visualization results are shown in the column labeled 4096 in Fig.~\ref{fig:compare}. We can observe that  ASFM-Net can not only predict the missing part of the object but also preserve the details of the input point cloud. For example, in the fifth rows of Fig.~\ref{fig:compare}, TopNet and PCN are totally failed. They cannot complete the missing areas, even destroy the original shape of the lamp. RFA attempts to repair the lamp but fails to output a satisfactory result. In contrast, the completed point cloud by the proposed ASFM-Net preserves the detailed structure from the input and reconstructs the missing lamp cover successfully. 


\subsection{Evaluation on Completion3D benchmark}\label{sec:evaluation on Completion3D benchmark}
Completion3D benchmark\protect\footnotemark[1] is released by TopNet~\cite{topnet}, which is a subset of the ShapeNet dataset derived from the PCN dataset. Different from the PCN dataset, the resolution of both partial and complete point clouds is 2048 points. Moreover, each complete model is only corresponding to one partial point cloud. The train/test split is the same as the PCN dataset.
\footnotetext[1]{\url{https://completion3d.stanford.edu/}.}

In the column labeled 2048 in Fig.~\ref{fig:compare}, we present the results of the visual comparison between ASFM-Net and other approaches, from which we can observe the more reasonable ability to infer the missing parts and the more effective fidelity of ASFM-Net. The quantitive comparison results of ASFM-Net with the other state-of-the-art point cloud completion approaches are shown in Table~\ref{table:c3d on c3d}. It is apparent from this table that the proposed ASFM-Net achieves the best performance concerning chamfer distance averaged across all categories. Compared with the second-best method VRC-Net~\cite{vrcnet}, ASFM-Net improves the performance of averaged chamfer distance with a margin of $ 17.7\%$. 

\subsection{Car completion on KITTI dataset}\label{sec:car completion on kitti dataset}
Following PCN, we evaluate the proposed method for car completion on the KITTI~\cite{kitti} dataset. The KITTI dataset includes many real-scanned partial cars collected by Velodyne 3D laser scanner. In this experiment, we take one sequence of raw scans. It contains 2483 partial car point clouds, which are collected from 98 real cars under 425 different frames. Same as PCN, every point cloud is completed with a model trained on cars from ShapeNet, and then transformed back to the world frame. Notably, there are no ground truth point clouds in the KITTI dataset.

\begin{table}[!htbp]
	\centering
	\renewcommand{\arraystretch}{1.0}
	\setlength{\tabcolsep}{0.9mm}
	\caption{ Fidelity error (FD) and consistency comparison on the KITTI  dataset.}
	\label{table:kitti consistency}
	\begin{tabular}{c|c|c|c|c|c}
		\hline
		Methods          & Input & TopNet~\cite{topnet} & PCN~\cite{pcn}   & RFA~\cite{sfa}   & ASFM-Net \\ \hline
		FD &- & 0.041  & 0.041 & 0.072 & \textbf{0.025}   \\ \hline
		Consistency  & 0.052 & \textbf{0.014}  & 0.016 & 0.033 & 0.020  \\ \hline
	\end{tabular}
\end{table}

\begin{figure}[!htbp]
	\centering
	\includegraphics[width=0.44\textwidth]{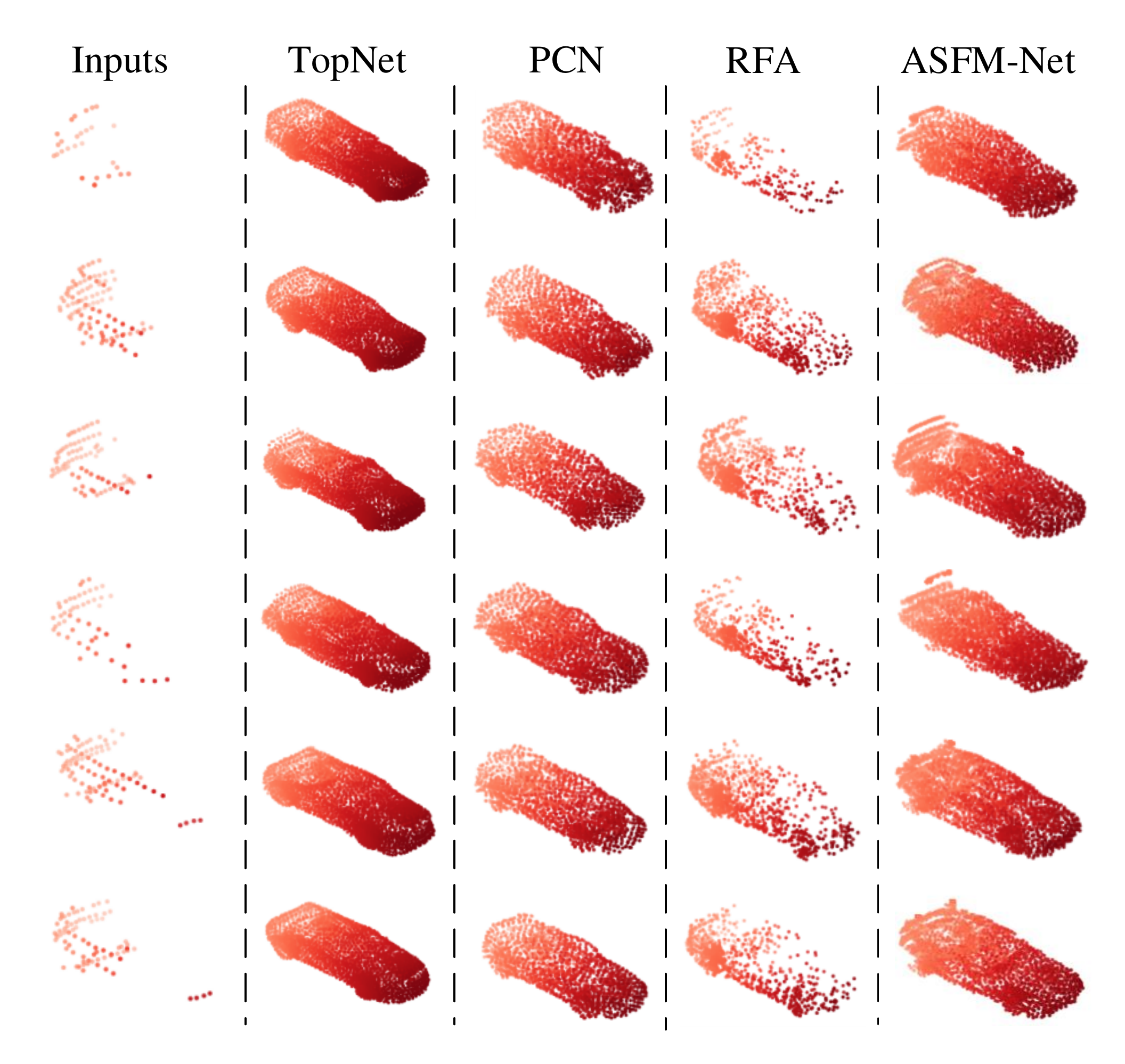}
	\caption{Qualitative comparison on the KITTI dataset. From left to right: Raw point clouds of the same vehicle scanned in consecutive frames, shape complete ion based on TopNet~\cite{topnet}, PCN~\cite{pcn}, RFA~\cite{sfa} and our method ASFM-Net.}
	\label{fig:kitti_oneframe}
\end{figure}

Therefore, fidelity error (FD), consistency, and minimal matching distance (MMD) are proposed by PCN as evaluation metrics. 
MMD is to measure how much the completion output resembles a typical car. However, we think this metric is not meaningful since we only trained on the car category. In other words, we already know the prior information that this object must be a car. Thus, we ignore MMD as an efficient evaluation metric in this paper.
The fidelity is to measure the similarity between the input and completed point clouds, which calculates the average distance between each point in the input and the nearest point in the completed point cloud. Consistency is to measure how consistent the outputs reconstructed by networks are against variations in the inputs, which calculates the average Charmfer distance among the completion results of the same instance in consecutive frames.
The quantitative results are shown in Table~\ref{table:kitti consistency}. 
From Table~\ref{table:kitti consistency}, it can be seen that ASFM-Net achieves excellent fidelity performance. The value is far below the other three methods, which illustrates ASFM-Net can preserve the original shape of the input effectively. 
The consistency performance of ASFM-Net is slightly weaker than TopNet and PCN. However, ASFM-Net improves the performance by a large margin compared with raw inputs. To explore the consistency performance more clearly, we visualize the results completed by different methods for one specific vehicle in different frames, as shown in Fig.~\ref{fig:kitti_oneframe}. The raw point clouds of the same vehicle have significantly different shape appearances since they are collected at different moments from a moving vehicle. This is the reason that the value of consistency for inputs is high. From Fig.~\ref{fig:kitti_oneframe}, we can see that PCN and TopNet are good at generating a general shape of the car while our method ASFM-Net reconstructs plausible shapes and keeps fine-grained details from the input. It demonstrates that ASFM-Net is more flexible when the appearance of the input point cloud changes greatly. 

\section{Discussion}
\subsection{Ablation Study}\label{sec:ablation study}

The ablation study evaluates the effectiveness of different proposed modules in our network, including both the pre-trained asymmetrical Siamese auto-encoder and refinement unit. All experiments are conducted on the Completion3D benchmark. The CD-P is selected as the evaluation metric.

\begin{table*}[ht]
	\centering
	\renewcommand{\arraystretch}{1.0}
	\setlength{\tabcolsep}{3.45mm}
	\caption{Ablation studies of asymmetrical Siamese auto-encoder and refinement unit on the Completion3D benchmark.}
	\label{table:c3d on c3d ablation}
	\begin{tabular}{c|c c c c c c c c|c}
		\hline
		\multirow{2}{*}{Methods} & \multicolumn{9}{c}{Chamfer Distance($ 10^{-3} $)}                                       \\ \cline{2-10} 
		& Airplane   & Cabinet   & Car    & Chair   & Lamp    & Sofa    & Table   & Watercraft   & Average   \\ \hline \hline
		PCN~\cite{pcn}           & 14.10          & 27.03          & 20.28          & 27.02          & 24.99          & 27.16          & 22.30          & 17.03          & 22.49          \\
		SA-PCN Decoder     & 13.32          & 25.69          & 19.82          & 23.67          & 22.98          & 24.12          & 22.46          & 16.84          & 21.11          \\
		PCN-Refine    & 10.57          & 23.48          & 18.87          & 21.45          & 18.37          & 22.59          & 18.27          & 13.98          & 18.45          \\
		TopNet~\cite{topnet}        & 13.88          & 28.07          & 19.69          & 24.74          & 23.36          & 26.12          & 22.48          & 16.66          & 21.88          \\
		SA-TopNet Decoder  & 13.09          & 25.55          & 20.22          & 24.37          & 23.03          & 24.78          & 21.55          & 17.42          & 21.25          \\
		TopNet-Refine & 10.96          & 23.86          & 18.78          & 21.41          & 17.85          & 22.05          & 18.31          & 14.08          & 18.41          \\
		w/o Refine    & 16.02          & 27.32          & 20.58          & 28.38          & 26.08          & 27.74          & 24.78          & 17.56          & 23.56          \\
		ASFM-Net       & \textbf{10.39} & \textbf{22.42} & \textbf{18.24} & \textbf{19.32} & \textbf{17.31} & \textbf{21.66} & \textbf{17.82} & \textbf{13.74} & \textbf{17.61}
		\\ \hline
	\end{tabular}
\end{table*}
\begin{table*}[!htbp]
	\centering
	\renewcommand{\arraystretch}{1.0}
	\setlength{\tabcolsep}{4mm}
	\caption{Quantitative  comparison on novel categories on the Completion3D benchmark. Point resolutions for the output and ground truth are 2048. For CD-P, lower is better.}
	\label{table:c3d on test novel}
	\begin{tabular}{c|c c c c c c c c|c}
		\hline
		\multirow{2}{*}{Methods} & \multicolumn{9}{c}{Chamfer Distance($ 10^{-3} $)}                                                  \\ \cline{2-10} 
		& bed      & bench     & bookshelf     & bus    & guitar     & motor    & pistol     & skateboard     & Average    \\ \hline \hline
		TopNet~\cite{topnet}              & 39.76 & 20.64 & 28.82 & 17.77 & 15.62 & 22.52 & 22.13 & 18.26 & 23.19 \\
		PCN~\cite{pcn}                 & 38.73 & 21.28 & 29.26 & 18.47 & 17.19 & 23.10 & 20.34 & 17.70 & 23.26 \\ 
		RFA~\cite{sfa}                 & 34.67 & 19.27 & 23.38 & 18.05 & 17.21 & 21.33 & 19.93 & 18.95 & 21.98 \\ \hline
		ASFM-Net              & \textbf{31.94} & \textbf{17.31} & \textbf{23.19}  & \textbf{17.02} & \textbf{11.97} &\textbf{16.81}  &\textbf{15.83}  & \textbf{14.50} & \textbf{18.57} \\ \hline
	\end{tabular}
\end{table*}
{\bf Pre-trained asymmetrical Siamese auto-encoder} In this section, we evaluate the effectiveness of the asymmetrical Siamese auto-encoder which learns shape prior to introduce a feature matching strategy. We use the pre-trained asymmetrical Siamese auto-encoder instead of the encoder modules of PCN and TopNet but keep their decoder modules respectively, referred as SA-PCN Decoder and SA-TopNet Decoder. The quantitative results are illustrated in Table~\ref{table:c3d on c3d ablation}. With the proposed asymmetrical Siamese auto-encoder module, our method brings significant improvements on the average chamfer distance by $ 6.1\% $ and 2.9\%, respectively, when compared with PCN and TopNet. This is due to the global features extracted by the asymmetrical Siamese auto-encoder include shape priors, while the global features directly encoded from the partial inputs are less informative. The comparison results demonstrate the superiority of the Siamese auto-encoder for 3D global feature learning for object completion.

{\bf The refinement unit} We choose the TopNet~\cite{topnet} and PCN~\cite{pcn} as the baseline. The refinement unit is first just integrated into TopNet and PCN, referred as TopNet-Refine and PCN-Refine. The quantitative results are shown in Table~\ref{table:c3d on c3d ablation}. Comparing with TopNet and PCN, both TopNet-Refine and PCN-Refine improve the performance across all categories significantly. The performance of TopNet-Refine and PCN-Refine also exceeds the average chamfer distance of TopNet and PCN by 15.86\% and 17.96\%, respectively. In addition, we remove the refinement unit from the proposed ASFM-Net and only asymmetrical Siamese auto-encoder is used, referred as w/o Refine. These results show that the refinement unit contributes to learning more perfect shape information with fine-grained details and is helpful for other point completion networks.


\subsection{Robustness Test}\label{sec:robustness test}
To further evaluate the robustness of the models, we conduct experiments on the input point clouds with various visible ratios. The visible ratio $R_v$ between the partial and the complete point clouds is defined as:
\begin{equation}
	{R_v} = {N_p}/{N_c}\label{f:rv}
\end{equation}
where $ N_p $ is the resolution of the points of a partial point cloud under the currently visible radius in the spherical space and $ N_c $ is the total points of complete point clouds. $R_v$ ranges from $ 20\%$ to $ 80\%$ with a step of 20\%.
\begin{table}[!htbp]
	\centering
	\renewcommand{\arraystretch}{1.0}
	\setlength{\tabcolsep}{1.0mm}
	\caption{Quantitative  comparison on known categories under different visible ratio. The CD-P is reported by TopNet, PCN, RFA and our method, multiplied by $10^{3}$. }
	\label{table:robust}
	\begin{tabular}{c|c c c c c c c}
		\hline
		\multirow{2}{*}{Methods} & \multicolumn{7}{c}{Visible Ratio}                             \\ \cline{2-8} 
		& 20\%  & 30\%  & 40\%  & 50\%  & 60\%  & 70\%  & 80\%  \\ \hline \hline
		TopNet~\cite{topnet}              & 41.92 & 34.72 & 30.16 & 27.27 & 25.44 & 24.27 & 23.47 \\ 
		PCN~\cite{pcn}                 & 45.65 & 37.82 & 33.08 & 30.06 & 27.64 & 25.77 & 24.01 \\ 
		RFA~\cite{sfa}                 & 41.58 & 34.61 & 29.64 & 25.83 & 22.74 & 20.74 & 19.06 \\ \hline
		ASFM-Net              & \textbf{39.94} & \textbf{31.71} & \textbf{25.97}  & \textbf{22.03} & \textbf{19.29} &\textbf{17.47}  &\textbf{16.27}   \\ \hline	
	\end{tabular}
\end{table}

\begin{figure}[!htbp]
	\centering
	\includegraphics[width=0.46\textwidth]{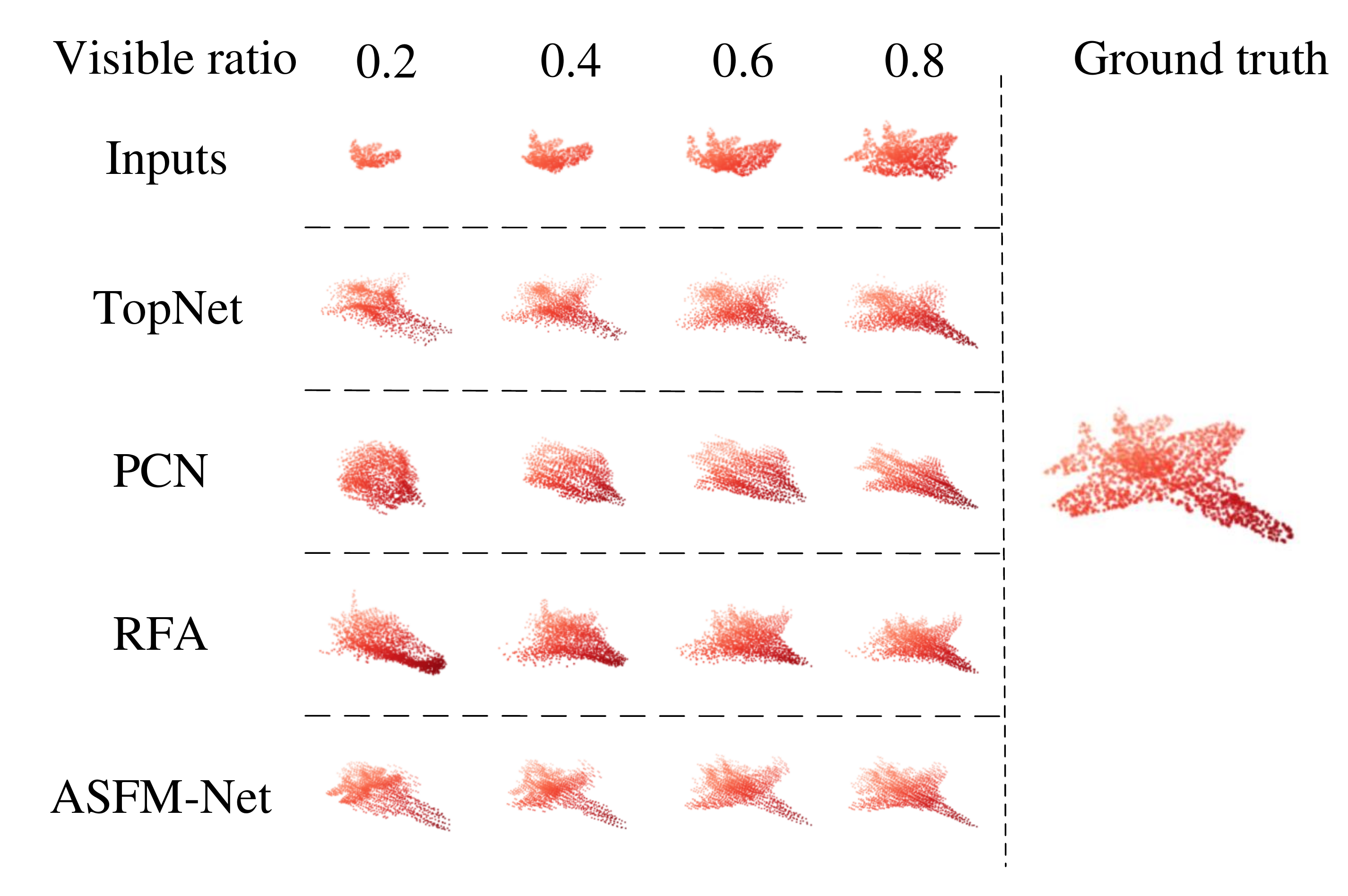}
	\caption{Qualitative comparison on inputs with different visible ratio. From top to down: Partial point clouds with different levels of visibility, completed point clouds by TopNet, PCN, RFA and our ASFM-Net.}
	\label{fig:robust}
\end{figure}

The quantitative and qualitative results are shown in Table~\ref{table:robust} and Fig.~\ref{fig:robust}, respectively. From them, we can conclude the following two conclusions: (1) Our method can deal with high missing degrees more robustly. Even though the visibility is only 0.2, ASFM-Net still generates the overall shape of the airplane. Both TopNet~\cite{topnet} and RFA~\cite{sfa} miss a good aircraft shape and generate uneven points. The result of PCN~\cite{pcn} is completely failed. (2) ASFM-Net reaches the best performance no matter in any visible ratios, which demonstrates our method is more robust to occlusion data.

\subsection{Completion on novel categories}\label{sec:completion on novel categories}

Since the asymmetrical Siamese auto-encoder is unsupervised trained using the models with known categories in the pre-built database, it is essential to explore its impact on the entire network ASFM-Net when facing unknown objects. In this section, we select eight novel categories for evaluation from the ShapeNet dataset, which are divided into two groups: one is the bed, bench, bookshelf, and bus (visually similar to the training categories), another is guitar, motorbike, pistol, and skateboard (visually dissimilar to the training categories). All experiments are conducted on the Completion3D benchmark. 

\begin{figure}[!htbp]
	\centering
	\includegraphics[width=0.45\textwidth]{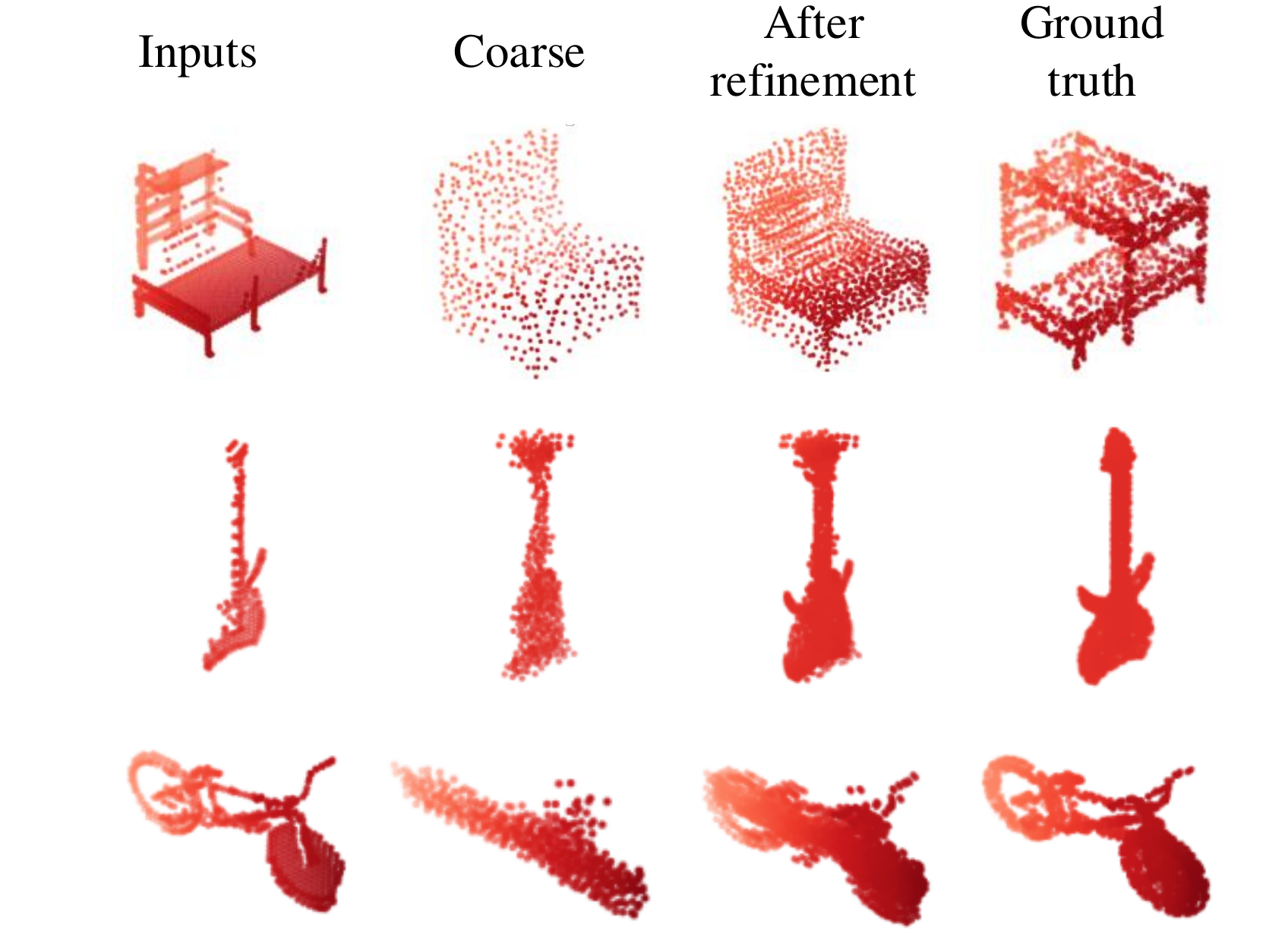}
	\caption{Qualitative point cloud completion result on the novel categories. 'Coarse' means the outputs only completed by the asymmetrical Siamese auto-encoder. 'After refinement' means the final point cloud completed by our ASFM-Net including a refinement unit. }
	\label{fig:noveltest}
\end{figure}

The qualitative and quantitative results are shown in Table~\ref{table:c3d on test novel} and Fig.~\ref{fig:noveltest}, respectively. From Fig.~\ref{fig:noveltest}, we can see the coarse outputs generated by the asymmetrical Siamese auto-encoder are wrong. It mistakenly completes the bed as the chair, the guitar as the lamp, and the motorbike as the watercraft. This is consistent with our expectation that the prior category information is learned from the known training categories. However, even if the asymmetrical Siamese auto-encoder provides wrong results, ASFM-Net can possibly reconstruct satisfactory point clouds (Row 2,3) thanks to the refinement unit. Besides, as shown in Table~\ref{table:c3d on test novel}, ASFM-Net outperforms other state-of-the-art methods on all novel categories. Notably, our method can improve performance by a large margin on visually dissimilar categories (e.g. the pistol and skateboard). This demonstrates that ASFM-Net has better generalizability than all previously tested state-of-the-art methods.

\section{Conclusion}
In this paper, we propose a novel end-to-end network ASFM-Net for point cloud completion. We design an asymmetrical Siamese auto-encoder, using a feature matching strategy to learn fruitful shape priors from complete point clouds in the pre-built database. Besides, we introduce the iterative refinement unit to preserve the information of inputs and reconstruct complete point clouds with find-grained details. Experiments show that our ASFM-Net achieves 1st place in Completion3D benchmark. According to discussion on experimental results, especially ablation studies, we can discover that the refinement unit contributes more to the performance of the ASFM-Net. In the future, we will explore the application of the completed results for 3D perceptual tasks, like instance extraction and type classification.

\section{Acknowledgement}
This research was supported by the China Scholarship Council. This work was supported by NSFC Grant No. 61901343, the National Defense Key Lab of China (6142411184413), the 111 Project (B08038), the Fundamental Research Funds for the Central Universities.

\bibliographystyle{ACM-Reference-Format}
\bibliography{ASFM-Net}

\end{document}